\documentclass[11pt,a4paper]{article}
\usepackage{times,latexsym,url}
\usepackage[T1]{fontenc}
\usepackage[acceptedWithA]{tacl2021v1}
\usepackage{
    multirow,
    placeins,
    booktabs,
    amsmath,
    amssymb,
    algorithm,
    algpseudocode,
    xcolor,
    array,
    graphicx
}

\title{
Embedding Consistency Regulation for\\
Persistent Semantic Routing in Compact Language Models
}

\author{
  Chung-Wei (Victor) Yuan \\
  Taipei, Taiwan \\
  \texttt{victor@yvic.dev}
}

\date{}

\begin{document}
\maketitle
\begin{abstract}
Language models are increasingly deployed under tight memory and compute constraints, where semantic routes (e.g., language, tone, or intent) may be assigned before decoding. 
However, these routes may not remain reflected in generated responses. 
We study this phenomenon as \emph{persistent semantic routing}.
We investigate this property using Embedding Consistency Regulation (ECR) as a minimal probe. 
ECR derives a short discrete prefix for each input using a frozen embedding model and applies the same prefix construction during both training and inference.
In controlled prefix-assignment experiments, ECR achieves 99.93\% $\pm$ 0.12\% language routing accuracy, while random, shuffled, and dummy controls remain near chance (34--36\%). 
This gap is not reliably captured by likelihood: some controls achieve competitive NLL despite losing routing consistency.
We introduce Response Routing Recoverability (RRR), an encoder-independent diagnostic that tests whether assigned routes remain observable after generation. 
Routing recoverability persists across multilingual settings, unsupervised partitions, reduced precision, and transfer tasks.
These findings establish routing persistence as a distinct and measurable property of language models, complementary to task performance and likelihood.
\end{abstract}

\section{Introduction}
\label{introduction}

Language models are increasingly being deployed directly on user devices, driven by concerns over privacy, security, latency, and regulations such as GDPR~\citep{Voigt2017GDPR}. Under severe memory and compute constraints, compact language models have become increasingly important~\citep{xu2024ondevice,liu2024mobilellm}. In such deployments, semantic routes such as language, tone, or intent are often assigned before decoding to guide model behavior.
Model compression~\citep{bucilua2006model} and knowledge distillation~\citep{hinton2015distilling} are common approaches for compact-model deployment. Many distillation methods focus on matching teacher outputs or representations~\citep{gou2021knowledge,sun2019patient,wang2020minilm}, rather than preserving semantic routing assignments specified before decoding.

Compact models may exhibit a distinct failure mode: routes assigned before decoding may not remain reflected in generated responses. Related phenomena have been observed as language drift in multilingual generation~\citep{li2026languagedrift} and persona or behavioral drift in dialogue systems~\citep{li2024measuring}. We refer to the preservation of pre-decoding semantic routes through generation as \emph{persistent semantic routing}, and to its failure as \emph{semantic routing drift}.
This observation raises two questions. 

First, can route persistence be studied without introducing additional learned controllers? 
We address this using Embedding Consistency Regulation (ECR), a minimal input-conditioning framework in which each input is mapped to a frozen embedding-space assignment and serialized as a deterministic discrete prefix shared across training and inference. 

Second, how should route persistence be measured after generation?
We answer this with Response Routing Recoverability (RRR), an encoder-independent diagnostic that measures whether assigned routes remain observable after generation using embedding spaces independent of the assignment process.
\paragraph{Contributions.}
\begin{itemize}
    \item We formulate \emph{persistent semantic routing}: whether a pre-decoding route remains reflected after generation.
    \vspace{10mm}
    \item We study this property with ECR, a deterministic prefix derived from a frozen embedding space.
    \item We introduce RRR, an encoder-independent diagnostic for measuring route persistence from generated responses.
    \item We show that routing persistence is measurable, reproducible, and not reliably captured by likelihood or output similarity.
\end{itemize}

\begin{figure*}[t]
    \centering
    \includegraphics[width=.99\textwidth]{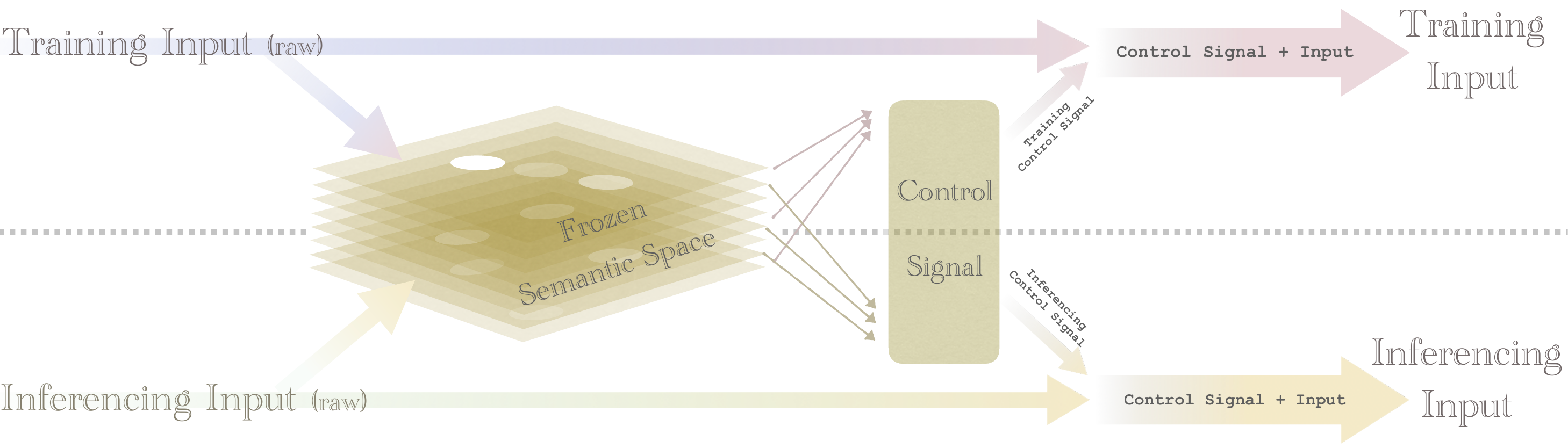}
    \vspace{-1mm}
    \caption{
    \textbf{Embedding Consistency Regulation (ECR).}
    Training and inference inputs are aligned through a shared semantic projection and control-prefix construction pipeline.
    The resulting prefix acts as a reusable semantic anchor for compact models under fixed-capacity deployment.
    }
    \label{fig:ecr}
\end{figure*}

\vspace{3mm}
\section{Background and Related Work}
\subsection{Semantic Routing}
LLM systems frequently assign inputs to specific routes before generation, including multilingual routing~\citep{bandarkar2026multilingual}, retrieval routing~\citep{zhang2025ragrouter}, tool selection~\citep{schick2023toolformer}, agent workflows~\citep{yao2022react,khattab2024dspy}, and task-specific processing~\citep{huang2025routereval,srivatsa2024harnessing}.
Routing is also used to balance accuracy, latency, and computational cost~\citep{chen2024frugalgpt,ong2025routellm}.
While mixture-of-experts models perform routing internally through expert selection~\citep{shazeer2017moe,fedus2021switch}, the routing considered in this work is assigned externally at the semantic or deployment level.
Such systems therefore rely not only on selecting an appropriate route before decoding, but also on whether that assigned route remains reflected in the generated output.

\vspace{3mm}
\subsection{Semantic Routing Drift}
Language drift has been observed in multilingual generation, particularly when models operate across languages or incorporate retrieved context~\citep{li2026languagedrift}. 
In dialogue systems, persona and behavioral consistency have also been shown to degrade over extended interactions~\citep{li2024measuring}. 
Reduced precision during deployment can further shift model outputs, even when the generated text remains fluent~\citep{frantar2023optq,lin2025awq}. 
These observations suggest a more general problem: semantic constraints assigned before generation can gradually weaken or drift during decoding, particularly in compact models or when using reduced precision.

\subsection{Route Preservation and Recoverability}
Standard language model evaluation focuses primarily on output quality through task accuracy, likelihood, or semantic similarity~\citep{wang2024mmlu,thakur2021beir}. 
Related lines of work have shown that upstream information can leave recoverable traces in generated text. Prompt recovery and model inversion studies demonstrate that instructions can be partially reconstructed from outputs~\citep{zhang2024extractingprompts,gao2024dory}, while watermarking methods show that signals introduced during generation can remain detectable afterward~\citep{kirchenbauer2023watermark,hou2023semstamp,hou2024ksemstamp,dabiriaghdam2025simmark}. 

These works suggest that information introduced before or during generation may persist in model outputs. However, prior research has mainly focused on recovering prompts, instructions, training data, or detecting watermark signatures. 
In contrast, this work examines whether \emph{structured semantic routing assignments} remain recoverable from generated responses.

\vspace{3mm}
\subsection{Semantic Conditioning Mechanisms}
A range of methods influence model behavior by modifying the input. These include discrete control codes such as CTRL~\citep{keskar2019ctrl}, continuous prefix tuning~\citep{li2021prefixtuning}, prompt tuning~\citep{lester2021power}, and guided decoding approaches such as PPLM~\citep{Dathathri2020Plug}, GeDi~\citep{krause2021gedi}, FUDGE~\citep{yang2021fudge}, and DExperts~\citep{liu2021dexperts}. Most of these methods aim to steer output attributes or adapt the model toward desired behaviors.

The objective of this work is different. Prompt tuning and prefix tuning learn conditioning signals by optimizing prefix parameters for task objectives. ECR, by contrast, derives a route from a frozen embedding space for each input, serializes it as a short discrete prefix, and introduces no additional trainable prefix parameters. This design keeps the assigned route external and explicitly nameable, making post-generation recoverability a well-defined question. Accordingly, the focus here is on studying routing persistence rather than the effects of learned conditioning capacity. We evaluate this persistence under controlled perturbations, including random, shuffled, dummy, and unsupervised assignments.
\section{Embedding Consistency Regulation}

ECR is an input conditioning framework that applies the same semantic projection and prefix-construction procedure during both training and inference.
We use \emph{regulation} to denote persistent control of this semantic conditioning pathway, rather than an explicit regularization loss on model parameters or latent states.
\subsection{Semantic Coordinate System}
\label{sec:ecr_coord}

As illustrated in Fig.~\ref{fig:ecr}, ECR uses a frozen embedding model as a fixed semantic space from which routing assignments are derived.
The coordinate system contains two factor groups:
(\textit{i}) an \textbf{output language control variable} $L$ and
(\textit{ii}) a set of \textbf{anchoring manifolds} $\mathcal{M}_{\text{anc}}$,
instantiated in our experiments as \textbf{emotion} and \textbf{intent}.
For each anchoring manifold
$m\in\mathcal{M}_{\text{anc}}$
with label set $\mathcal{Y}_m$,
we precompute a prototype vector
$p_{m,y}$
for each label
$y\in\mathcal{Y}_m$.

\subsection{Projection and Discretized Control Codes}
\label{sec:ecr_proj_disc}

Given a query $Q$, we compute a normalized embedding $\tilde e = g(Q)/\|g(Q)\|$. For each anchoring manifold $m\in\mathcal{M}_{\text{anc}}$ and label $y\in\mathcal{Y}_m$, we compute the affinity score

\[
s_{m,y}=\cos(\tilde e,\tilde p_{m,y}),
\]

where $\tilde p_{m,y}$ denotes the normalized prototype vector associated with label $y$. The highest-affinity labels are selected to form the routing assignment

\[
\hat Y_m=\text{Top-}k\{s_{m,y}\},
\]

where $k$ controls the number of retained labels per manifold. In our experiments, we retain the top-2 labels for each anchoring manifold. The resulting routing assignments are serialized into a discrete control prefix

\[
T=[\tau(L)] \Vert [\tau(m,\hat y)]_{m\in\mathcal{M}_{\text{anc}},\,\hat y\in\hat Y_m},
\]

where $\tau(\cdot)$ maps semantic assignments to discrete control tokens. The prefix $T$ is prepended to the original query and shared across both training and inference.

\subsection{Training and Inference}

Given training pairs $(Q,A)$, we construct the prefix $T$ (Alg.~\ref{alg:ecr_prefix}) and optimize the standard supervised objective $\theta \leftarrow \arg\min_{\theta}\mathcal{L}(f_\theta([T,Q]),A)$ on the conditioned input $[T,Q]$. 
At inference, the same prefix construction procedure is applied and decoding is performed on $[T,Q]$.

\begin{algorithm}[ht]
\caption{ECR prefix construction}
\label{alg:ecr_prefix}
\small
\begin{algorithmic}[1]
\Require Query $Q$, prototypes $\{p_{m,y}\}$, token map $\tau(\cdot)$, top-$k$, threshold $\gamma$
\Ensure Prefix $T$
\State $e \gets g(Q)$
\State $\tilde e \gets e/\|e\|$
\State $T \gets [\tau(L)]$
\For{$m \in \mathcal{M}_{\text{anc}}$}
    \For{$y \in \mathcal{Y}_m$}
        \State $s_{m,y}\gets \cos(\tilde e,\tilde p_{m,y})$
    \EndFor
    \State $\hat Y_m \gets \text{Top-}k\{y:s_{m,y}\ge\gamma\}$
    \If{$|\hat Y_m|<k$}
        \State pad with neutral labels
    \EndIf
    \State $T \gets T \Vert [\tau(m,\hat y)]_{\hat y\in\hat Y_m}$
\EndFor
\State \Return $T$
\end{algorithmic}
\end{algorithm}

\begin{figure*}[!t]
    \centering
    \includegraphics[width=\textwidth]{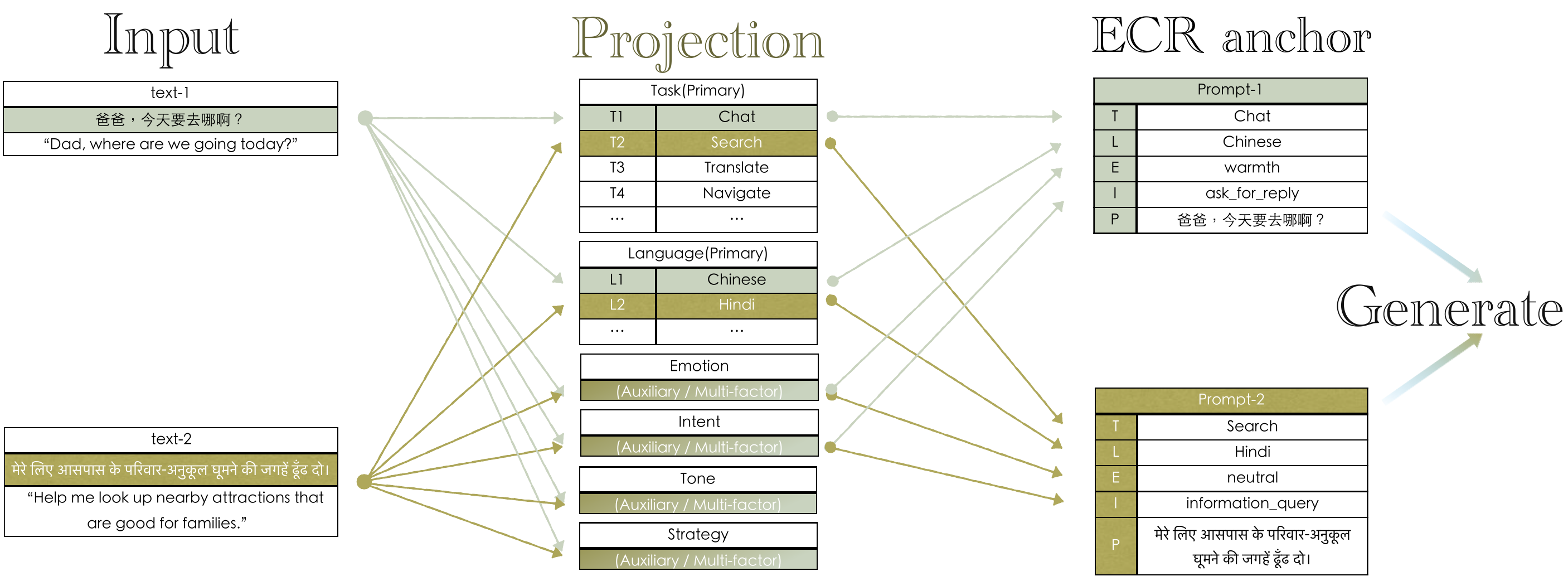}
    \vspace{-3mm}
    \caption{
    \textbf{Overview of ECR.}
    Queries are projected into a fixed semantic space and serialized into discrete control prefixes shared across training and inference.
    }
    \label{fig:semantic_projection_pipeline}
\end{figure*}

\subsection{Practical Instantiation}
\label{sec:impl_details}

As shown in Fig.~\ref{fig:semantic_projection_pipeline}, ECR constructs prefixes by matching query embeddings against fixed label prototypes rather than retrieved documents.
Prefix construction requires one frozen embedding call, cosine comparisons against prototypes, and token lookup; decoding is unchanged except for the short prefix.
In the main multilingual experiments, we instantiate ECR with one output-language slot and four semantic-anchor slots:
\[
T=[\tau(L),\tau(E_1),\tau(E_2),\tau(I_1),\tau(I_2)].
\]

Here, $L\in\{\text{EN},\text{ZH},\text{HI}\}$ denotes the target answer language (English, Chinese, and Hindi), while $E_1,E_2$ and $I_1,I_2$ denote the top-2 emotion and intent anchors.
We use threshold $\gamma=0.7$ and neutral default identifiers when fewer than two labels pass the threshold.
Control identifiers are serialized as compact string prefixes and prepended to the query.

\section{Experimental Setup}
\label{sec:experiments}

We evaluate ECR on compact multilingual models using the same training procedure across all comparisons.
A 3B model trained without ECR is included as a larger reference.

\subsection{Model and Training Configuration}
\label{sec:training_setup}
We evaluate 1B and 3B decoder-only Transformer models compatible with the Llama-3.2 family~\cite{llama3}.
Models are trained under both full precision (FP32) and reduced precision (BF16).
The 1B model is used for all SFT-only and SFT+ECR comparisons, while the 3B model is trained without ECR.
For all experiments, the frozen semantic coordinate system is instantiated using BGE-M3~\cite{bge}. Emotion and intent prototypes are constructed from curated label descriptions encoded by the frozen model, following the procedure described in Section~\ref{sec:ecr_coord}.

ECR uses the standard supervised training objective and does not introduce any additional routing network, learned controller, or continuous prefix parameters.
All models use the same tokenization, optimization settings, initialization, batching strategy, and train/test splits unless otherwise specified.

\subsection{Dataset}

\label{sec:data_stats}

To evaluate whether an assigned route remains recoverable after generation, we require a corpus with known routing labels and a controlled generation process. Real-world instruction corpora rarely provide such conditions because semantic factors are typically unlabeled and mixed sources may introduce confounding artifacts.

We therefore generate a multilingual instruction--response dataset using a standardized pipeline. Reference responses are produced by a large language model with aligned semantic structure across English, Chinese, and Hindi. All samples are generated automatically without manual filtering or post-editing, ensuring that routing assignments are known by construction.

This controlled setup enables direct evaluation of routing persistence. Construction details and release specifications are provided in Appendix~\ref{app:dataset_schema}.
\section{ECR with Semantic Labels}
\label{sec:eval_results}

We first characterize generation quality, then evaluate route preservation directly through prefix assignment controls and Response Routing Recoverability (RRR).

\subsection{Generation Diagnostics Across Metrics}

We report likelihood, output similarity, and representation geometry as complementary diagnostics of generation behavior.
Representative training trajectories are given in Appendix~\ref{app:training_stability}.

\begin{table}[!t]
\centering
\resizebox{\columnwidth}{!}{
\begin{tabular}{p{0.4\linewidth} p{0.18\linewidth} p{0.18\linewidth} p{0.18\linewidth}}
\hline
Model & English & Chinese & Hindi \\
\hline
3B BF16 & 2.94 & 3.04 & 2.30 \\
\textbf{3B FP32} & \textbf{2.07} & \textbf{2.08} & \textbf{1.59} \\
1B BF16 & 3.14 & 3.33 & 2.49 \\
1B BF16 + ECR & 3.44 & 3.59 & 2.75 \\
1B FP32 & 3.59 & 3.50 & 2.76 \\
\textbf{1B FP32 + ECR} & \textbf{2.76} & \textbf{2.79} & \textbf{2.12} \\
\hline
\end{tabular}
}
\vspace{-3mm}
\caption{
Negative log-likelihood (NLL).
}
\label{tab:nll_results}
\end{table}

\begin{table}[!t]
\centering
\resizebox{\columnwidth}{!}{
\begin{tabular}{p{0.34\linewidth} p{0.14\linewidth} p{0.14\linewidth} p{0.18\linewidth} p{0.22\linewidth}}
\hline
Model & E5 & LaBSE & mMPNet & mMiniLM \\
\hline
3B BF16        & 0.8455 & 0.6794 & 0.5871 & 0.5423 \\
\textbf{3B FP32}& \textbf{0.8647} & \textbf{0.7315} & \textbf{0.6479} & \textbf{0.6122} \\
1B BF16        & 0.8353 & 0.6569 & 0.5296 & 0.4866 \\
1B BF16 + ECR  & 0.8371 & 0.6637 & 0.5223 & 0.4835 \\
1B FP32        & 0.8372 & 0.6707 & 0.5259 & 0.5010 \\
\textbf{1B FP32 + ECR} & \textbf{0.8558} & \textbf{0.7091} & \textbf{0.6078} & \textbf{0.5772} \\
\hline
\end{tabular}
}
\vspace{-1mm}
\caption{
Output embedding similarity (OSim) between reference answers $A$ and generated outputs $\hat{A}$.
}
\label{tab:osim_results}
\end{table}

The 3B FP32 model achieves the lowest NLL (Table~\ref{tab:nll_results}), confirming that capacity still matters.
More relevant to ECR, however, the NLL behavior differs across precisions.
Under FP32, ECR reduces the 1B NLL from 3.59 to 2.76.
Under BF16, where training is already more stable, ECR instead introduces a small NLL penalty.
This suggests that NLL reflects optimization behavior in addition to generation quality, and we therefore treat it as a secondary diagnostic rather than evidence of route persistence.

A similar precision dependence appears in output similarity (Table~\ref{tab:osim_results}).
OSim is measured only with encoders independent of the conditioning model:
E5~\cite{e5} and three Sentence Transformer encoders~\cite{reimers-2020-sentence-bert}---LaBSE, mMPNet, and mMiniLM.
Under BF16, ECR produces small and inconsistent changes across encoders, whereas under FP32, improvements appear consistently across all four.
Like NLL, OSim reflects generation quality rather than route persistence.

Table~\ref{tab:geom_results} provides a complementary internal view of representation organization over the full input--target sequence.
Intra and Inter refer to the average Euclidean distances between samples sharing the same route and samples with different routes, respectively. Ratio is defined as Intra divided by Inter.
ECR reduces both Intra and Ratio under both precisions, indicating tighter organization among samples sharing the same route.
However, this remains an internal characterization of the trained model.
Whether the assigned route remains observable after generation is evaluated independently through RRR, discussed next.
\begin{table}[!t]
\centering
\resizebox{\columnwidth}{!}{
\begin{tabular}{p{0.35\linewidth} p{0.2\linewidth} p{0.2\linewidth} p{0.2\linewidth}}
\hline
Model & Intra $\downarrow$ & Inter& Ratio $\downarrow$ \\
\hline
3B BF16 & 42.51 & 43.54 & 0.976 \\
\textbf{3B FP32} & \textbf{39.20} & \textbf{41.05} & \textbf{0.955} \\
1B BF16 & 48.02 & 48.95 & 0.981 \\
1B BF16 + ECR & 41.17 & 42.48 & 0.969 \\
1B FP32 & 46.51 & 47.33 & 0.983 \\
\textbf{1B FP32 + ECR} & \textbf{39.66} & \textbf{41.91} & \textbf{0.946} \\
\hline
\end{tabular}
}
\vspace{-3mm}
\caption{Representation geometry metrics.}
\label{tab:geom_results}
\end{table}

\begin{table}[!t]
\centering
\resizebox{\columnwidth}{!}{
\begin{tabular}{p{0.18\linewidth} p{0.35\linewidth} p{0.08\linewidth} p{0.34\linewidth}}
\hline 
Run & Lang. Acc. & NLL  & OSim(E5)  \\
\hline
ECR       & \textbf{0.9993 $\pm$ 0.0012} & 6.12 & \textbf{0.897 $\pm$ 0.0003} \\
Random    & 0.3430 $\pm$ 0.0030 & \textbf{6.11} & 0.858 $\pm$ 0.0003 \\
Shuffled  & 0.3640 $\pm$ 0.0080 & 6.11 & 0.858 $\pm$ 0.0004 \\
Dummy     & 0.3553 $\pm$ 0.0059 & 6.14 & 0.856 $\pm$ 0.0014 \\
\hline
\end{tabular}
}
\vspace{-2mm}
\caption{
Prefix assignment controls averaged over three seeds under 1B FP32.
}
\label{tab:random_control}
\end{table}

\begin{figure*}[t]
\centering
\includegraphics[width=.95\textwidth]{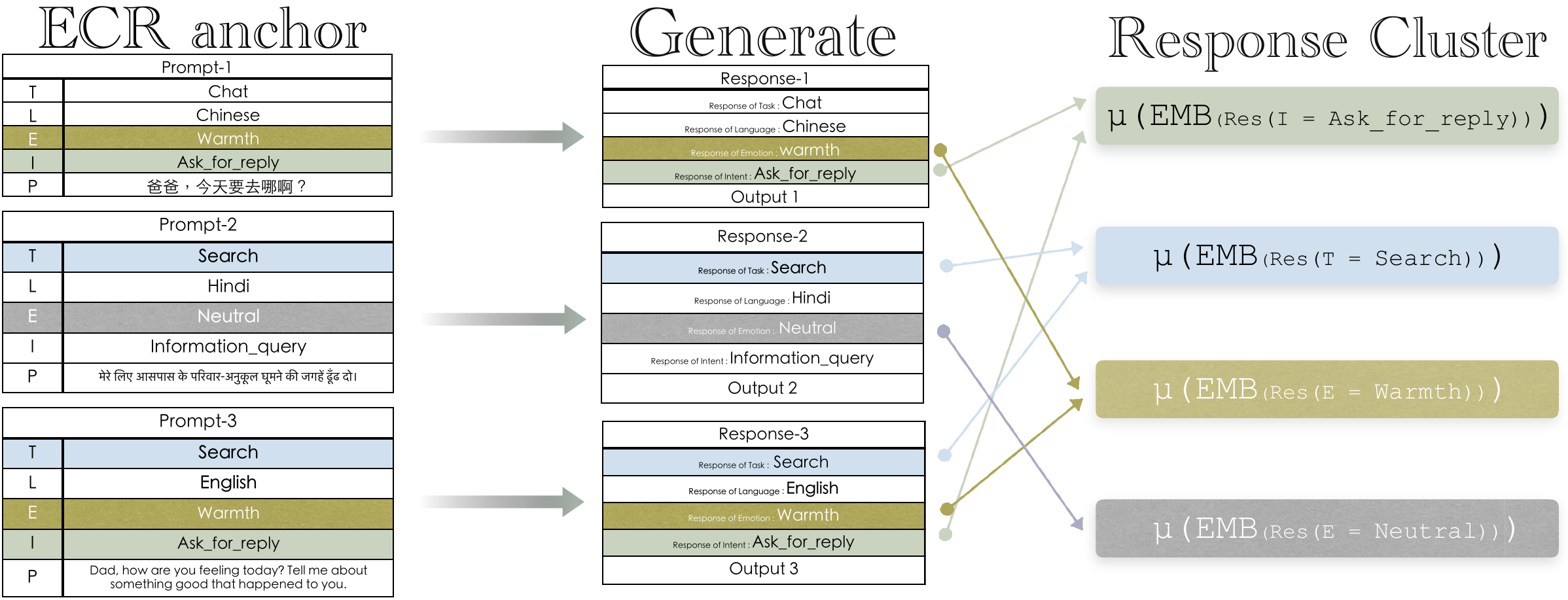}
\vspace{-2mm}
\caption{
\textbf{Response Routing Recoverability (RRR).}
Generated responses are embedded with an external encoder and compared to response centroids for each routing slot.
RRR measures whether the originally assigned route ranks among the nearest centroids after generation.
}
\label{fig:response_cluster}
\end{figure*}

\subsection{Prefix Assignment Controls}
\label{sec:random}

To test whether strong routing recoverability comes from meaningful semantic assignment rather than simply from adding extra prefix tokens, we train three control conditions on the 20k multilingual subset. All three controls use the same prefix length and token budget as ECR, and all results are averaged over three random seeds.

\begin{itemize}
    \item \textbf{Random}: Prefix labels are assigned independently of the input, breaking the connection between input semantics and route.
    \item \textbf{Shuffled}: The overall label distribution is preserved, but the specific label for each example is randomly permuted.
    \item \textbf{Dummy}: Every example receives the same fixed prefix, adding tokens without any input-dependent routing signal.
\end{itemize}

As shown in Table~\ref{tab:random_control}, language routing clearly separates ECR from the controls. 
Random, shuffled, and dummy prefixes all remain near the three-language chance level ($\approx 33\%$) under script-based detection. In contrast, ECR achieves near-perfect language control (99.93\% $\pm$ 0.12\%) across all seeds.
This demonstrates that neither the existence of prefix tokens nor the global label distribution is sufficient; effective routing requires a consistent semantic alignment between the input and its assigned prefix.

Interestingly, NLL does not reflect this difference.
Random and shuffled achieve slightly lower NLL than ECR despite completely losing language routing ability.
In contrast, ECR achieves the highest semantic similarity measured by OSim while maintaining almost perfect routing accuracy.

\subsection{Response Routing Recoverability}
\label{sec:RRR}
Language routing is relatively easy to check manually, since a generated response is usually clearly English, Chinese, or Hindi. ECR, however, assigns more than just language. 
For emotion and intent, the key question is whether a response assigned to a specific route — such as joy, concern, or advice — remains aligned with the reference responses that belong to that same route. 
This cannot be reduced to fluency, correct language, or overall similarity to the reference answer.

NLL and OSim do not directly measure routing persistence. 
NLL evaluates token-level likelihood against the reference, while OSim measures overall embedding similarity. Both can stay favorable even when the assigned emotion or intent has drifted. 
RRR addresses this by checking whether each assigned routing slot can still be recovered from the generated response itself.

\subsubsection{Recovery Procedure}

Each ECR prefix contains five routing slots: one language, two emotion, and two intent assignments.
RRR evaluates these slots using sentence encoders independent from the encoder used to construct ECR.
For each external encoder and each routing slot, we construct one centroid per label value from reference-answer embeddings.

Generated responses are then embedded by the same external encoder and ranked against the corresponding slot-wise centroids.
RRR@$K$ counts a slot as recovered if its original assigned label appears among the top-$K$ nearest centroids.
Scores are averaged across examples and four external sentence encoders (E5, LaBSE, mMPNet, and mMiniLM) to reduce dependence on any single embedding space.
Fig.~\ref{fig:response_cluster} illustrates this recovery procedure, and Algorithm~\ref{alg:routing_recovery} summarizes it.

\begin{algorithm}[h]
\caption{Encoder-agnostic recovery of ECR routing assignments from generated responses}
\label{alg:routing_recovery}
\small
\begin{algorithmic}[1]
\Require
$\{(Q_i,A_i)\}_{i=1}^{N}$,
$\{\hat A_i\}_{i=1}^{N}$,
$\{T_i\}_{i=1}^{N}$,
encoder $h$,
top-$K$
\Ensure
$\{RRR@K_j\}$
\State Parse routing labels
$Y_{i,j}$,
$j\in\{L,E_1,E_2,I_1,I_2\}$
\For{$j$}
\State Construct centroids
$\tilde\mu_{j,y}$
\State Embed generations
$\tilde z_i$
\State Retrieve top-$K$
labels
$\widehat{\mathcal Y}_{i,j}^{K}$
\State
$RRR@K_j
\leftarrow
\frac1N
\sum_i
\mathbf1[
Y_{i,j}
\in
\widehat{\mathcal Y}_{i,j}^{K}
]$
\EndFor
\State Return
$\{RRR@K_j\}$
\end{algorithmic}
\end{algorithm}

\begin{table*}[t]
\centering
\resizebox{\textwidth}{!}{
\begin{tabular}{lcccccccc}
\toprule
& \multicolumn{4}{c}{RRR@1}
& \multicolumn{4}{c}{RRR@2} \\
\cmidrule(lr){2-5}
\cmidrule(lr){6-9}
Metric
& Teacher
& ECR
& Random
& Shuffled
& Teacher
& ECR
& Random
& Shuffled \\
\midrule
L
& 86.85
& 92.07 ± 0.13
& 34.22 ± 1.08
& 32.17 ± 0.29
& 97.03
& 98.25 ± 0.11
& 66.40 ± 1.09
& 65.56 ± 0.36 \\
E1
& 14.75
& 13.58 ± 0.45
& 4.80 ± 0.28
& 1.96 ± 0.23
& 28.50
& 25.41 ± 0.62
& 9.88 ± 0.61
& 6.29 ± 0.06 \\
E2
& 12.80
& 11.18 ± 0.18
& 4.93 ± 0.14
& 2.09 ± 0.18
& 23.47
& 20.22 ± 0.18
& 10.07 ± 0.53
& 6.13 ± 0.26 \\
I1
& 19.23
& 19.94 ± 0.64
& 3.44 ± 0.35
& 0.57 ± 0.16
& 32.42
& 32.45 ± 0.45
& 6.77 ± 0.24
& 2.02 ± 0.15 \\
I2
& 10.65
& 7.65 ± 0.30
& 3.49 ± 0.25
& 0.61 ± 0.10
& 19.95
& 14.96 ± 0.39
& 7.15 ± 0.43
& 2.63 ± 0.27 \\
\midrule
L\_E1
& 13.20
& 12.31 ± 0.34
& 1.87 ± 0.04
& 0.55 ± 0.12
& 27.62
& 24.82 ± 0.55
& 7.00 ± 0.51
& 4.13 ± 0.15 \\
L\_E2
& 11.05
& 10.22 ± 0.05
& 1.73 ± 0.38
& 0.55 ± 0.13
& 22.90
& 19.90 ± 0.24
& 6.63 ± 0.47
& 3.89 ± 0.06 \\
L\_I1
& 16.95
& 18.42 ± 0.60
& 1.28 ± 0.28
& 0.17 ± 0.06
& 31.40
& 31.72 ± 0.42
& 4.68 ± 0.23
& 1.24 ± 0.16 \\
L\_I2
& 9.28
& 6.92 ± 0.35
& 1.45 ± 0.13
& 0.18 ± 0.13
& 19.38
& 14.68 ± 0.42
& 4.88 ± 0.39
& 1.53 ± 0.25 \\
\bottomrule
\end{tabular}
}
\vspace{2mm}
\caption{
Response Routing Recoverability (RRR@K) averaged over three random seeds and four external embedding models (E5, LaBSE, mMPNet, and mMiniLM).
}
\label{tab:rrr}
\end{table*}

\subsubsection{Recoverability Results}
RRR measures whether the route assigned before decoding can still be recovered from the generated response by matching against slot-wise centroids in an external embedding space. 
RRR@$K$ considers a route recovered if the correct label appears among the top-$K$ nearest centroids.

Unlike standard evaluation settings, the Teacher responses do not serve as a perfect upper bound. 
Because the multilingual QA pairs are taken directly from the aligned corpus without manual filtering or correction, they naturally contain some cross-lingual noise. 
A script-based check reveals English text in 13.08\% of the Hindi reference responses, which partly explains why even the Teacher ceiling is imperfect.

Despite this, ECR stays close to the Teacher across most routing factors. On average, ECR preserves roughly 90\% of the Teacher's recoverable routing structure while clearly outperforming both random and shuffled controls. 
The advantage is especially clear for language routing: ECR reaches 92.07\% RRR@1 and 98.25\% RRR@2, approaching or even matching the Teacher under the same protocol, whereas random and shuffled assignments remain near chance. 

Emotion and intent routing are more difficult, and composed factors (e.g., L\_E1) are harder still because multiple dimensions must be preserved simultaneously. 
Even so, ECR remains well above the random and shuffled controls. This indicates that generated responses preserve structured semantic routing beyond language matching alone.
\section{ECR with Unsupervised Partitioning}
The results in Section~\ref{sec:eval_results} are based on manually defined semantic factors (language, emotion, and intent), with supervision applied only to the answer tokens. A natural question is whether the observed routing recoverability depends on this particular supervised setup, or whether it also emerges under a different training regime. If recoverability mainly requires semantically interpretable factors, then removing the hand-defined taxonomy should weaken the effect. On the other hand, if recoverability primarily depends on the existence of a stable partitioning structure, then explicit semantic meaning may not be necessary.
\subsection{Experimental Setup}
\label{sec:uns_dataset}
Apart from full-sequence supervision, this experiment differs from the main setting \emph{only in the dataset and label source.}
All other settings follow Section~\ref{sec:training_setup}.
Models are trained for two epochs, and cluster identifiers are registered as atomic control tokens before training.
Additional details are provided in Appendix~\ref{app:uns_partition_details}.

\paragraph{Dataset.}
To isolate semantic partitioning from language identity, we reuse the dataset introduced in Section~\ref{sec:data_stats} and retain only the English QA pairs.
Only the query and answer fields are used in this setting.

\paragraph{Source of labels.}
Queries are projected into a frozen BGE-M3 embedding space and partitioned using K-means clustering.
Given a normalized query embedding $\tilde e = g(Q)/\|g(Q)\|$, each query is assigned to a cluster $c(Q)\in\{1,\dots,C\}$, whose identifier is converted into a discrete control token $\tau(c(Q))$ prepended to the input sequence.

We evaluate partition granularities $C\in\{1,4,16,32,64,256\}$.
The goal is not to identify an optimal clustering level, but to examine whether routing behavior remains recoverable when manually defined semantic labels are replaced by increasingly fine-grained unsupervised partitions.
The overall prefix-construction procedure is summarized in Algorithm~\ref{alg:uns_ecr_prefix}.

\begin{algorithm}[ht]
\caption{Unsupervised prefix construction}
\label{alg:uns_ecr_prefix}
\footnotesize
\begin{algorithmic}[1]
\Require Query $Q$; embedding model $g(\cdot)$; $C$-cluster partitioner $\pi_C$; token map $\tau(\cdot)$
\Ensure Control prefix $T$
\State $\tilde e \gets g(Q)/\|g(Q)\|$
\State $c(Q) \gets \pi_C(\tilde e)$
\State $T \gets [\tau(c(Q))]$
\State \Return $T$
\end{algorithmic}
\end{algorithm}

\subsection{Results under Unsupervised Partitioning}
\label{sec:uns_result}
When we replace the curated factors with K-means cluster identifiers, the main pattern still holds (Table~\ref{tab:uns_results}). Routes remain recoverable well above chance under both FP32 and BF16, even though likelihood does not reliably improve. Under FP32, moderate partition counts reduce NLL relative to the single-cluster baseline, with the best result at $  C{=}32  $, though the effect is not monotonic. Under BF16, the NLL advantage disappears completely.

RRR shows a clearer and more consistent picture. Recoverability stays substantially above the corresponding $1/C$ chance level across all partition granularities and both precisions, including settings where NLL shows no benefit. 
As partitions become finer, absolute recoverability decreases, but the chance baseline drops even faster. At $  C{=}256  $, ECR still achieves 11.94\% recoverability against a chance rate of only 0.39\% — roughly thirty times higher. 
The corresponding Teacher recoverability averages 53.48\% across the four external encoders, suggesting that routing information is partially preserved through generation but not completely.

Overall, these results align with the main experiments: routing induced purely through frozen embedding-space partitions remains recoverable from generated responses, even in the absence of manually defined semantic taxonomies.

\begin{table}[!t]
\centering
\small
\resizebox{\columnwidth}{!}{
\begin{tabular}{@{}llccccc@{}}
\toprule
& & & & \multicolumn{3}{c}{RRR@1} \\
\cmidrule(lr){5-7}
Prec. & Partition & NLL & OSim & Teacher & ECR & Chance \\
\midrule
\multirow{3}{*}{BF16}
& $C{=}1$ (SFT) & 5.623 & 0.9030 & -- & -- & -- \\
& $C{=}32$      & 5.642 & 0.9035 & 47.38 & 32.24 & 3.13 \\
& $C{=}256$     & 5.652 & \textbf{0.9038} & 53.48 & 23.51 & 0.39 \\
\midrule
\multirow{6}{*}{FP32}
& $C{=}1$ (SFT) & 2.738 & 0.8746 & -- & -- & -- \\
& $C{=}4$       & 2.747 & 0.8751 & 70.23 & 34.74 & 25.00 \\
& $C{=}16$      & 2.695 & 0.8770 & 52.08 & 24.86 & 6.25 \\
& $C{=}32$      & 2.653 & \textbf{0.8777} & 47.38 & 21.68 & 3.13 \\
& $C{=}64$      & 2.798 & 0.8738 & 45.35 & 11.98 & 1.56 \\
& $C{=}256$     & 2.695 & 0.8772 & 53.48 & 11.94 & 0.39 \\
\bottomrule
\end{tabular}
}
\vspace{-4mm}
\caption{
Unsupervised partitioning results.
RRR@1 is averaged across four external encoders.
Chance denotes random recovery ($1/C$).
}
\label{tab:uns_results}
\end{table}

\begin{table}[!t]
\centering
\resizebox{\columnwidth}{!}{
\begin{tabular}{p{0.3\linewidth} p{0.2\linewidth} p{0.18\linewidth} p{0.18\linewidth} p{0.18\linewidth}}
\toprule
Backend & Format & $C{=}4$ & $C{=}32$ & $C{=}256$ \\
\midrule
\multirow{2}{*}{HuggingFace}
& FP32 & 34.74 & 21.68 & 11.94 \\
& FP16 & 34.49 & 21.70 & 11.73 \\
\midrule
\multirow{4}{*}{GGUF}
& Q6 & 28.48 & 29.54 & 16.23 \\
& Q5 & 32.42 & 28.88 & 15.13 \\
& Q4 & 48.51 & 20.42 & 13.29 \\
& Q3 & 38.21 & 11.36 & 6.84 \\
\midrule
& Chance & 25.00 & 3.13 & 0.39 \\
\bottomrule
\end{tabular}
}
\vspace{-3mm}
\caption{
RRR@1 across deployment formats derived from the same FP32-trained ECR checkpoint.
}
\label{tab:uns_precision_rrr}
\end{table}

\subsection{Evaluation Across Deployment Formats}
Although the above results use an FP32-trained checkpoint, practical deployment typically relies on reduced precision formats such as FP16 or GGUF quantization.

Using the FP32 ECR checkpoint from the unsupervised partitioning experiments (Section~\ref{sec:uns_result}), we export the model in several formats---native FP32 and FP16, plus GGUF Q6--Q3 via llama.cpp~\citep{llamacpp2023}---and recompute RRR@1 (Table~\ref{tab:uns_precision_rrr}). Routes remain recoverable well above chance in every format and at every granularity. At $C{=}32$, recoverability ranges from 11.4\% to 29.5\% against a 3.13\% baseline; even at the finer $C{=}256$ setting, it ranges from 6.8\% to 16.2\% against a 0.39\% baseline.

FP32 and FP16 perform nearly identically, while the GGUF variants show non-monotonic behavior across bit widths---Q4, for example, sometimes recovers better than Q6. Because all formats are derived from the same checkpoint, these differences are not caused by training variations. However, they do not cleanly separate the effect of quantization from other format-specific implementation and runtime factors, so we treat them as an empirical observation rather than evidence that any particular format preserves routing better.

The more robust conclusion is that routing information largely survives deployment-oriented conversion, while the precise effect of each format on routing remains an open question.

\begin{table*}[t]
\centering
\resizebox{\textwidth}{!}{
\begin{tabular}{lccccc ccc}
\toprule
&&&
\multicolumn{3}{c}{RRR@1}
&
\multicolumn{3}{c}{RRR@2}
\\
\cmidrule(lr){4-6}
\cmidrule(lr){7-9}
Partition
& NLL
& OSim(E5)
& Teacher
& ECR
& Random
& Teacher
& ECR
& Random \\
\midrule

$C{=}1$ 
& 3.611 $\pm$ 0.128
& 0.806 $\pm$ 0.003
& --
& --
& --
& --
& --
& -- \\

$C{=}4$
& 3.566 $\pm$ 0.071
& \textbf{0.811 $\pm$ 0.007}
& 31.65
& 20.22 $\pm$ 0.57
& 25.00
& 61.05
& 56.15 $\pm$ 0.58
& 50.00 \\

$C{=}8$
& 3.645 $\pm$ 0.051
& 0.809 $\pm$ 0.005
& 24.25
& 13.53 $\pm$ 0.73
& 12.50
& 41.75
& 29.65 $\pm$ 0.61
& 25.00 \\

$C{=}16$
& \textbf{3.521 $\pm$ 0.039}
& 0.800 $\pm$ 0.007
& 23.75
& 10.27 $\pm$ 0.93
& 6.25
& 38.15
& 19.40 $\pm$ 1.35
& 12.50 \\

$C{=}32$
& 3.531 $\pm$ 0.031
& 0.796 $\pm$ 0.005
& 25.65
& 6.30 $\pm$ 0.41
& 3.12
& 37.40
& 12.67 $\pm$ 1.11
& 6.25 \\

\bottomrule
\end{tabular}
}
\vspace{-4mm}
\caption{
FinQA transfer under unsupervised semantic partitioning.
Results are averaged over three random seeds and four external embedding models: E5, LaBSE, mMPNet, and mMiniLM.
}
\label{tab:finqa_transfer}
\end{table*}

\subsection{Structured Financial QA Transfer}

To test whether the partitioning effect generalizes beyond the synthetic multilingual corpus, we apply unsupervised partitioning to FinQA~\cite{chen2021finqa}, a numerical reasoning benchmark based on real financial reports. 
Its naturally occurring question--table pairs and expert-annotated answers provide a more challenging, externally sourced evaluation of whether the effect survives a domain shift.

We cluster the frozen embeddings of the combined question--table context, convert each cluster into an ECR prefix, and predict the execution answer. 
Table~\ref{tab:finqa_transfer} reports the transfer results under unsupervised partitioning, averaged over three random seeds. 
Random denotes the theoretical chance recovery rate ($1/C$).

Moderate partition counts reduce validation NLL relative to the single-cluster baseline (best at $C{=}16$), though the improvement is again non-monotonic. 
This pattern remains consistent across three seeds, with relatively small variance in both NLL and semantic similarity.

Routes remain recoverable above chance for $C \ge 8$. At $C{=}32$, ECR achieves 6.30\% RRR@1 compared with 3.12\% random recovery, and 12.67\% RRR@2 compared with 6.25\% random recovery. 
Recoverability decreases as the number of partitions increases, since finer-grained routing becomes harder to preserve.

The signal is weaker than in the multilingual setting, likely because FinQA answers are short and mostly numerical, leaving less textual evidence of the assigned route. 
Nevertheless, ECR consistently outperforms the random baselines. 
This suggests that the effect is not restricted to hand-designed taxonomies or tightly controlled conversational data.
Even when semantic partitions are derived directly from a real task, they can still give rise to measurable and stable routing behavior.

\section{Discussion}

\paragraph{Routing persistence is a separate axis.}
Task quality and route preservation are distinct objectives. 
A response can be fluent and task-appropriate while still weakening the route assigned before decoding. 
Conversely, a route can remain recoverable even when task performance stays the same. 
RRR specifically targets this second axis. It checks whether an assigned route is still reflected in the generated response, rather than recovering missing metadata. 
Using encoders independent of the assignment encoder is important because it ensures recoverability is not measured only inside the original conditioning space. 
In deployment, this distinction matters: if routing information disappears during generation, the logged assignment may no longer accurately describe the actual behavior of the response.

\paragraph{When ECR applies.}
ECR is intended for structured settings where the conditioning signal is built consistently at both training and inference, and where responses contain enough semantic content for the route to leave a recoverable trace. 
RRR itself serves as a practical check: when reference responses recover poorly under a given routing scheme, that scheme is not well reflected in the output space, so ECR is unlikely to help much. 
FinQA is the clearest example. Its targets are mostly short numerical answers, so recoverability remains limited even though the underlying task is structured. 
FinQA therefore extends the analysis to a structured QA domain where recoverability is naturally constrained by short outputs. 
Whether comparable routing persistence appears in open-ended dialogue remains an open question.

\paragraph{Precision and quantization.}
Routing persistence and likelihood behave differently under reduced precision. 
In both the main experiments and the unsupervised partitioning experiments, routes remain more recoverable than the random baselines, even in settings where NLL shows no improvement.
A similar pattern appears across quantized formats: even when likelihood changes are small or inconsistent, recoverability remains clearly above the random baselines.
These results suggest that changes in generation quality do not necessarily imply a corresponding loss of routing information, and that likelihood and routing persistence should be evaluated as separate dimensions.

\paragraph{Deployment.}
Because ECR operates through input conditioning alone, it can be applied to quantized, offline inference.
Appendix~\ref{app:system_overview} presents an end-to-end iOS deployment called \emph{Luna}.
Luna serves as a deployment case study demonstrating that ECR can operate in a practical on-device system.
In Luna, one routing field can be used to trigger local retrieval when needed. The controlled evaluation of ECR, however, comes from the prefix-assignment controls and RRR experiments, which do not involve retrieval.

\section{Future Work}
This work focuses on single-turn conditioning. 
Multi-turn interaction is a natural next challenge. 
Once the model starts conditioning on its own previous outputs, the initial route may gradually weaken. Periodic re-assignment---such as anchor refresh, lightweight state rewriting, or retrieval refresh between turns---may help maintain route persistence over longer conversations.

Another open question concerns which design choices matter most. Factors such as tokenization, how control identifiers are formed, and how supervision is composed may all influence whether a route remains recoverable after generation. Understanding these factors would help clarify how far ECR generalizes beyond the settings examined in this paper.

\section{Conclusion}
This work examines \textit{persistent semantic routing} — the property that a route assigned before decoding remains reflected in the generated response. We introduce Embedding Consistency Regulation (ECR) as a minimal input-conditioning framework: each input is mapped to a discrete prefix from a frozen embedding space and applied consistently during both training and inference.
Across multilingual, unsupervised partitioning, reduced precision, and deployment settings, assigned routes remain recoverable well above random, shuffled, and dummy controls. Likelihood does not reliably reflect this property; routing information can persist even when token-level loss shows no improvement. These results indicate that routing persistence is a distinct deployment consideration that should be evaluated alongside task performance and likelihood.
\bibliography{reference}
\bibliographystyle{acl_natbib}

\clearpage
\appendix
\noindent\textbf{Appendix overview.}
This appendix provides supplementary implementation details,
dataset construction procedures, prompt serialization,
and deployment specifications supporting the main results.

\section{Training and Implementation Details}
\label{app:model_training_details}

\paragraph{Training configuration.}
All models are trained using AdamW with learning rate $1\times10^{-6}$, linear decay, 200 warmup steps, $\beta_1=0.9$, $\beta_2=0.95$, $\epsilon=10^{-8}$, and weight decay $0.1$. Training uses a per-device batch size of 10 without gradient accumulation. FP32 runs use standard attention by default, while BF16 runs additionally use FlashAttention-2 and gradient checkpointing. Gradient clipping is disabled for all experiments (\texttt{max\_grad\_norm = 0}).

\paragraph{Conditioning and supervision.}
ECR prefixes are prepended before the query, and supervision is applied only to answer tokens. Aside from the use of ECR prefixes, all tokenization, optimization, batching, initialization, and train/test splits are shared across the corresponding baseline and ECR settings.

\section{Dataset Schema}
\label{app:dataset_schema}

Table~\ref{tab:data_stats} summarizes the multilingual corpus used in the main experiments.
The training corpus is derived from an aligned multilingual dataset (\texttt{raw\_data}) generated using GPT-5.
Each record contains English, Chinese, and Hindi query--answer pairs sharing the same emotion and intent annotations.
QA training instances are obtained by sampling a monolingual $(Q,A)$ pair while inheriting the associated semantic labels.
Table~\ref{tab:raw-data-schema} summarizes the raw-data schema.

\paragraph{Processed QA construction.}
For reproducibility, we release the processed QA pairs used for model training;
Table~\ref{tab:processed-data-schema} lists the released fields.
Preprocessing from \texttt{raw\_data} to QA training pairs includes multilingual text normalization and light OCR-style text corruption for robustness.
Query and answer languages are randomly selected per instance, while emotion and intent labels are inherited from the source record.
All models use a shared tokenizer.

For each training pair $(Q,A)$, the corresponding semantic labels are serialized into a control prefix and prepended to the input:
\[
[\tau(L),\tau(E_1),\tau(E_2),\tau(I_1),\tau(I_2)]
\ \Vert\
\texttt{Q:}\ Q
\ \Vert\
\texttt{A:}\ A.
\]
All identifiers are registered as atomic special tokens.
An example serialized input is shown below:

\begin{quote}
\small
\texttt{(h,E,A,G,L) \textbackslash n Q: Hey, what’s your name? \textbackslash n A: Hey! You can just call me your helpful assistant....}
\end{quote}

\begin{table}[!t]
\centering
\setlength{\tabcolsep}{2pt}
\resizebox{\columnwidth}{!}{
\begin{tabular}{p{0.23\linewidth} p{0.12\linewidth} p{0.12\linewidth} p{0.32\linewidth}}
\toprule
Language & Train & Test & Avg. Length \\
\midrule
English & 30k & 3.4k & 124.6 \\
Chinese & 30k & 3.3k & 138.0 \\
Hindi   & 30k & 3.3k & 172.2 \\
\midrule
Total   & 90k & 10k & 144.9 \\
\bottomrule
\end{tabular}
}
\caption{Dataset statistics.}
\label{tab:data_stats}
\end{table}

\begin{table}[!t]
\centering
\setlength{\tabcolsep}{6pt}
\resizebox{\columnwidth}{!}{
\begin{tabular}{p{0.08\linewidth} p{0.8\linewidth}}
\toprule
\textbf{Field} & \textbf{Description} \\
\midrule
L  & Language of the \emph{answer} $A$ \\
E1 & Primary emotion label ($e_1$) \\
E2 & Secondary emotion label ($e_2$) \\
I1 & Primary intent label ($i_1$) \\
I2 & Secondary intent label ($i_2$) \\
Q  & Query text (single-turn instruction) \\
A  & Reference answer text (supervised target) \\
\bottomrule
\end{tabular}
}
\caption{Schema of the processed QA training set.}
\label{tab:processed-data-schema}
\end{table}

\begin{table}[!t]
\centering
\resizebox{\columnwidth}{!}{
\begin{tabular}{p{0.23\linewidth} p{0.5\linewidth}}
\toprule
\textbf{Field} & \textbf{Description} \\
\midrule
dialog\_id & Dialogue identifier \\
e1 & Primary emotion label \\
e2 & Secondary emotion label \\
i1 & Primary intent label \\
i2 & Secondary intent label \\
EN\_Q & English query \\
EN\_A & English answer \\
ZH\_Q & Chinese query \\
ZH\_A & Chinese answer \\
HI\_Q & Hindi query \\
HI\_A & Hindi answer \\
\bottomrule
\end{tabular}
}
\caption{
Field definitions for records in the raw aligned data.
}
\label{tab:raw-data-schema}
\end{table}

\begin{figure*}[!t]
    \centering
    \includegraphics[width=\textwidth]{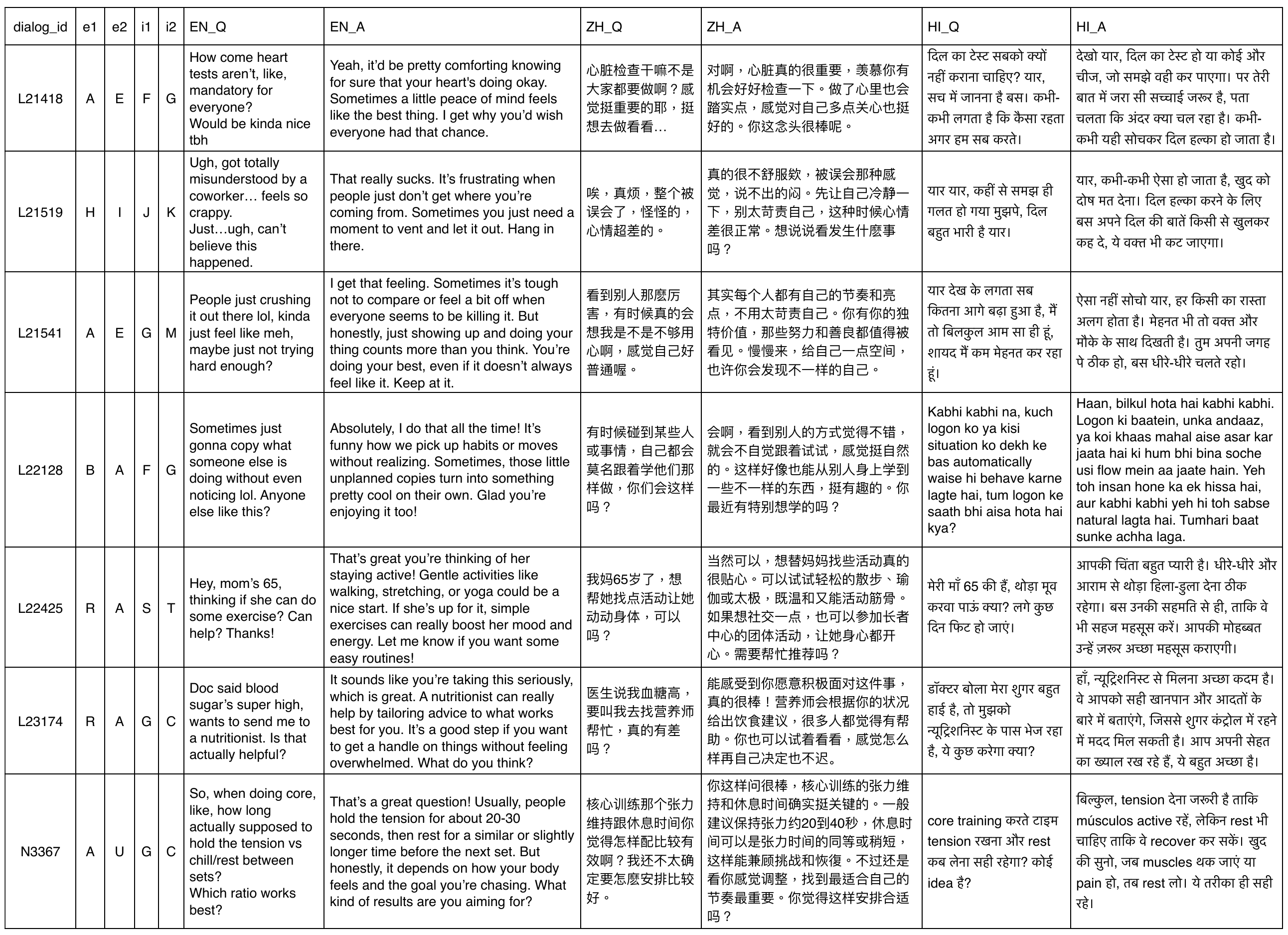}
    \vspace{-3mm}
    \caption{
    Example records from the aligned multilingual corpus (\texttt{raw\_data}).
    Each row contains English, Chinese, and Hindi query--answer pairs with shared emotion and intent labels, illustrating the cross-lingual alignment used to construct QA training pairs.
    }
    \label{fig:raw_data_examples}
\end{figure*}

\begin{figure*}[!h]
    \centering
    \includegraphics[width=\linewidth]{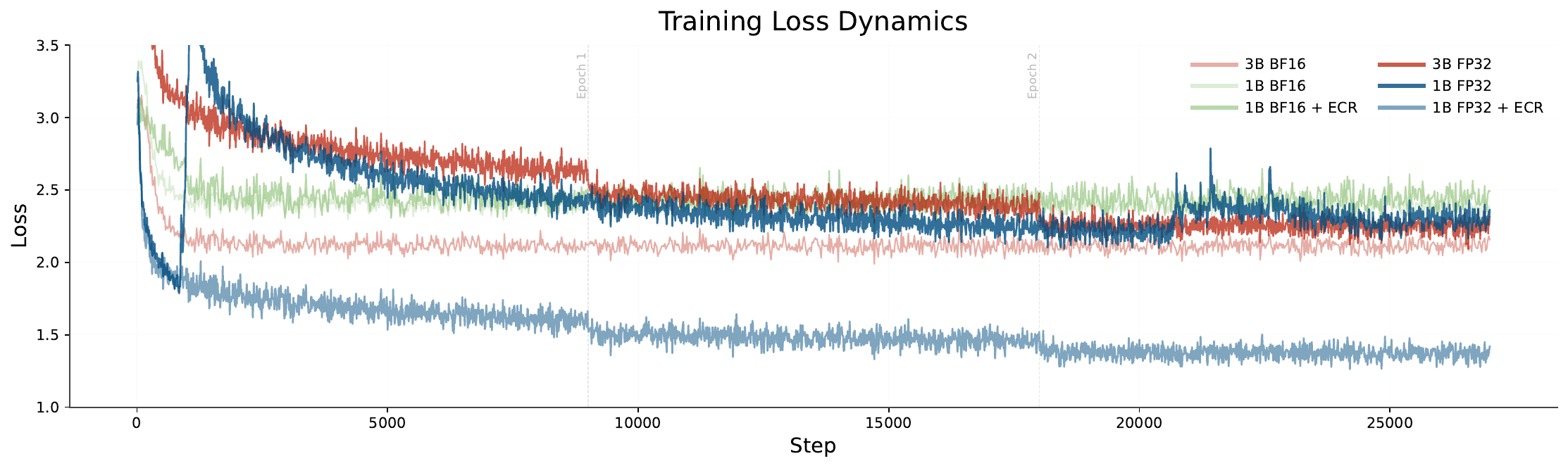}
    \vspace{-2mm}
    \caption{
    Representative training trajectories under the no--gradient-clipping setting.
    The curves illustrate optimization behavior within the evaluated runs rather than strict cross-configuration loss comparability.
    }
    \label{fig:loss_curve}
\end{figure*}

\section{Training Stability}
\label{app:training_stability}

Fig.~\ref{fig:loss_curve} illustrates representative training trajectories under the no--gradient-clipping setting.
The curves show that likelihood behavior is sensitive to training precision and optimization dynamics in the compact-model setting.
In the evaluated full-precision runs, the 1B baseline exhibits sharp early loss spikes, whereas 1B+ECR remains comparatively stable during training.

These observations help explain why NLL trends in Table~\ref{tab:nll_results} should not be read as direct evidence of route preservation.
They further motivate evaluating routing persistence through dedicated diagnostics such as RRR and controlled routing experiments.

\section{Unsupervised Partitioning Details}
\label{app:uns_partition_details}

\paragraph{Dataset subset.}

For the unsupervised partitioning experiments, we reuse the multilingual corpus introduced in Section~\ref{sec:data_stats} and retain only the English QA pairs.
All manually curated semantic annotations (language, emotion, and intent) are removed before training.
The resulting subset contains 45,000 examples, split into 40,000 training instances and 5,000 validation instances.

\paragraph{Prompt construction.}

For each training pair $(Q,A)$, the cluster assignment is converted into a discrete control token and prepended to the input sequence.
The resulting prompt is formatted as

\[
[\tau(c(Q))]
\ \Vert\
\texttt{Q:}\ Q
\ \Vert\
\texttt{A:}\ A.
\]

To avoid tokenizer decomposition effects, all cluster identifiers are registered as atomic special tokens before training.

\paragraph{Example serialization.}

For example, under the $C=256$ partition setting, a query assigned to cluster 21 may produce an input of the following form:

\begin{quote}
\small
\texttt{C<C21> \textbackslash n Q: Hey, what’s your name? \textbackslash n A: Hey! You can just call me your helpful assistant....}
\end{quote}

This setup isolates the effect of externally induced semantic organization without relying on manually curated semantic taxonomies.

\section{On-device System Overview}
\label{app:system_overview}

Luna is an on-device system that uses ECR to demonstrate practical deployment of the proposed method.
It runs a compact language model entirely locally with a paired retrieval index.
The application bundle includes model weights, the retrieval index, and a lightweight runtime layer for fully offline execution.
Within Luna, ECR remains a pure prefix construction mechanism.
For the on-device demo, one prefix field (a task token) can optionally trigger local retrieval before decoding.
This routing policy is specific to the deployment system and extends the core ECR conditioning mechanism studied in the main experiments.

ECR training in this work uses full-precision training, while deployment precision may vary depending on device constraints and runtime requirements.
For the Luna deployment example, the FP32-trained model is compressed into GGUF-formatted\footnote{\texttt{llama.cpp}: \url{https://github.com/ggml-org/llama.cpp}} Q4\_KM weights for inference on resource-constrained devices.

Tables~\ref{tab:system-overview}--\ref{tab:retrieval-specs} summarize the resulting deployment configuration.
Table~\ref{tab:system-overview} gives the overall system specification, Table~\ref{tab:device-config} reports the measured runtime footprint, and Table~\ref{tab:retrieval-specs} describes the local retrieval subsystem.

\begin{table}[!t]
\centering
\setlength{\tabcolsep}{4pt}
\resizebox{\columnwidth}{!}{
\begin{tabular}{p{0.3\linewidth} p{0.75\linewidth}}
\toprule
\textbf{Item} & \textbf{Specification} \\
\midrule
Model name & Luna-0.6B \\
Package size & 693 MB \\
Training data & 875{,}596 dialogue fragments \\
Throughput & $\sim$15 tokens/s (iPhone) \\
Core abilities & summarization; translation; offline RAG; sentiment analysis \\
\bottomrule
\end{tabular}
}
\caption{Summary of the Luna on-device system.}
\label{tab:system-overview}
\end{table}

\subsection{Hardware and Runtime Configuration}

Table~\ref{tab:device-config} reports the hardware, backend, model format, memory footprint, and throughput used in the on-device run.

\begin{table}[!t]
\centering
\setlength{\tabcolsep}{6pt}
\resizebox{\columnwidth}{!}{
\begin{tabular}{p{0.38\linewidth} p{0.72\linewidth}}
\toprule
\textbf{Item} & \textbf{Specification} \\
\midrule
Device & iPhone 14 (iOS 26) \\
Backend & \texttt{llama.cpp} with Metal acceleration \\
Model format & GGUF (Q4\_KM) \\
Runtime footprint & 693 MB (model + RAG index + runtime) \\
Peak RAM usage & 1.27 GB \\
Throughput & 14--15 tokens/s (offline) \\
\bottomrule
\end{tabular}
}
\caption{On-device runtime configuration and resource usage.}
\label{tab:device-config}
\end{table}

\subsection{Local Retrieval Subsystem}

Table~\ref{tab:retrieval-specs} summarizes the fully local retrieval index used by Luna.
No external communication is required during retrieval or generation.

\begin{table}[!t]
\centering
\resizebox{\columnwidth}{!}{
\begin{tabular}{p{0.4\linewidth} p{0.6\linewidth}}
\toprule
Index Type & HNSW, 64-d \\
Index Size & 48.9 MB \\
RAG Records & 381{,}943 text entries \\
Query Latency & $<2$ ms/query \\
Top-$k$ & 5 \\
Similarity Metric & Cosine distance \\
\bottomrule
\end{tabular}
}
\caption{Local retrieval subsystem specifications.}
\label{tab:retrieval-specs}
\end{table}

\begin{figure*}[!t]
    \centering
    \includegraphics[width=\textwidth]{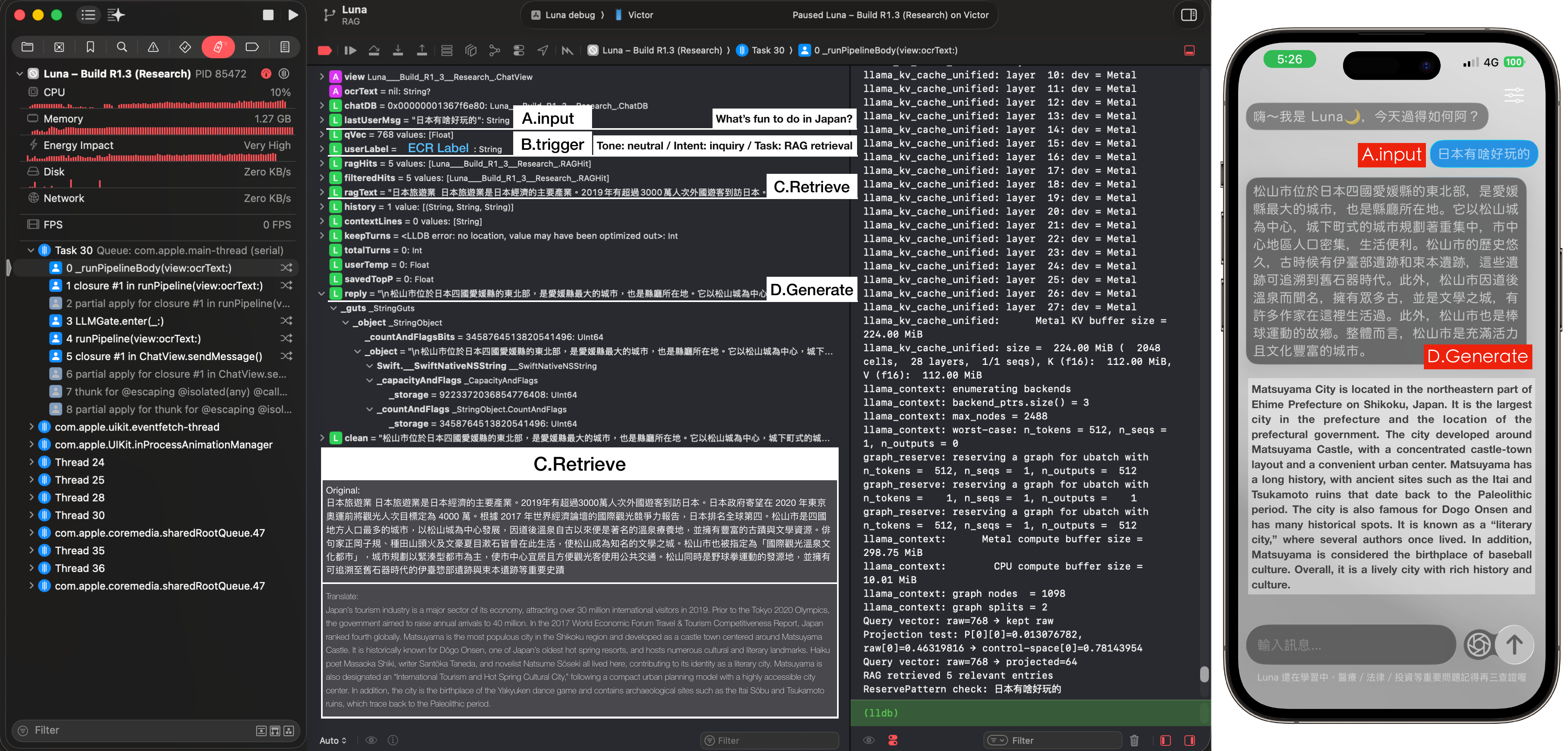}
    \vspace{-3mm}
    \caption{
    Runtime trace of the Luna on-device inference pipeline on iPhone~14.
    The trace illustrates ECR prefix construction, optional local retrieval, and quantized decoding through the llama.cpp runtime.
    }
    \label{fig:xcode-trace}
\end{figure*}

\subsection{End-to-End Execution Flow}
Fig.~\ref{fig:xcode-trace} illustrates the full inference pipeline 
running on an iPhone~14. For clarity, the trace is annotated with 
four stages that correspond to the ECR-aware execution path.

\textbf{(A) Input Normalization and Embedding.}
We normalized and tokenized user text with the shared multilingual tokenizer.
The tokens are then embedded to form the initial hidden states used by the model.

\textbf{(B) ECR Semantic Control Prefix.}
ECR projects the embedding onto its semantic anchors and retrieves the
corresponding control tokens as prefix.
The prefix contains the fields shown in the trace---Label,
Script, Tone (neutral), Intent (inquiry), and Task (RAG retrieve).
These values set up the initial embedding path.
They also control whether retrieval fires, which is why the trace branches.

\textbf{(C) Local Retrieval.}
In the example shown in Fig.~\ref{fig:xcode-trace}, the task is RAG retrieve. When the prefix signals retrieval,
the system runs a local vector-store search.
It returns the entries that fall in the same semantic region. The trace shows the projection to 64 dimensions,
cosine scoring, and HNSW top $K$ selection.

\textbf{(D) ECR-Aware Quantized Decoding.}
Decoding uses Metal-accelerated quantized weights with KV-cache reads, attention matvecs, command-buffer calls, and sampling shown in the trace.

\subsection{Case Study: Task Routing in Luna}

Fig.~\ref{fig:xcode-trace} illustrates how ECR is applied in the Luna on-device system. Given the limited capacity of a 0.6B model, the goal is not unrestricted open-domain generation but stable execution of lightweight tasks such as summarization.

In this example, the assigned task route triggers local retrieval, and the language model summarizes the retrieved context rather than answering from parametric memory alone. This case study demonstrates that ECR can support task-specific routing in a compact on-device setting, where controlled and predictable generation may be more practical than fully open-ended generation.

Luna is included as a deployment feasibility demonstration rather than a controlled evaluation of ECR in isolation. The controlled experiments that isolate ECR's contribution, without any retrieval component, are presented separately in Sec.~\ref{sec:random} and Sec.~\ref{sec:RRR}.

\clearpage
\iftaclpubformat
\fi
\end{document}